\documentclass{llncs}
\usepackage{amsmath}
\usepackage{algorithm}
\usepackage{algorithmic}
\usepackage{graphicx}
\usepackage{subfigure}
\usepackage{booktabs}

\newcommand{\set}[1]{\left\{#1\right\}}
\newcommand{\pr}[1]{\left(#1\right)}
\newcommand{\fpr}[1]{\mathopen{}\left(#1\right)}
\newcommand{\spr}[1]{\left[#1\right]}

\newcommand{\abs}[1]{{\left|#1\right|}}

\newcommand{\enset}[2]{\left\{#1 ,\ldots , #2\right\}}

\newcommand{\pp}{\textbf{PP}}
\newcommand{\np}{\textbf{NP}}

\newcommand{\define}{\leftarrow}

\newcommand{\iset}[2]{{#1} \cdots {#2}}
\newcommand{\freq}[1]{fr\fpr{#1}}

\newcommand{\ent}[1]{H\fpr{#1}}

\newcommand{\tree}[1]{\mathcal{#1}}
\newcommand{\cover}[1]{\mathcal{#1}}

\newcommand{\sep}[1]{S\fpr{#1}}
\newcommand{\vrt}[1]{V\fpr{#1}}
\newcommand{\edge}[1]{E\fpr{#1}}

\newcommand{\costmdl}[1]{C_{\text{MDL}}\fpr{#1}}

\newcommand{\dtname}[1]{\emph{#1}}

\title{Decomposable Families of Itemsets}
\author{Nikolaj Tatti and Hannes Heikinheimo}

\institute{HIIT Basic Research Unit, Department of Information and Computer Science,
Helsinki University of Technology, Finland,\\ \email{ntatti@cc.hut.fi} and \email{hannes.heikinheimo@tkk.fi}.}

\begin{document}
\maketitle
\begin{abstract}
The problem of selecting a small, yet high quality subset of patterns
from a larger collection of itemsets has recently attracted
lot of research. Here we discuss an approach to this problem using the notion
of decomposable families of itemsets. Such itemset families define a
probabilistic model for the data from which the original collection of
itemsets has been derived from. Furthermore, they induce a special
tree structure, called a junction tree, familiar from the theory of
Markov Random Fields. The method has several advantages. The junction
trees provide an intuitive representation of the mining results. From
the computational point of view, the model provides leverage for
problems that could be intractable using the entire collection of
itemsets. We provide an efficient algorithm to build decomposable
itemset families, and give an application example with frequency bound
querying using the model. Empirical results show that our algorithm
yields high quality results.
\end{abstract}

\section{Introduction}

Frequent itemset discovery has been a central research theme in the
data mining community ever since the idea was introduced by Agrawal
et. al \cite{AIS1993}. Over the years, scalability of the problem has
been the most studied aspect, and several very efficient algorithms
for finding all frequent itemsets have been introduced, Apriori
\cite{AMSTV1996} or FP-growth \cite{han2000} among others. However, it
has been argued recently that while efficiency of the mining task is
no longer a bottleneck, there is still a strong need for methods that
derive compact, yet high quality results with good application
properties \cite{han2006}.

In this study we propose the notion of decomposable families of
itemsets to address this need. The general idea is to build a
probabilistic model of a given dataset $D$ using a small well-chosen
subset of itemsets $\cover{G}$ from a given candidate family
$\cover{F}$. The candidate family $\cover{F}$ may be generated from
$D$ using some frequent itemset mining algorithm. A special aspect of
a decomposable family is that it induces a type of tree, called a
junction tree, a well-known concept from the theory of Markov Random
Fields~\cite{cowell99network}.

\begin{figure}[htb!]
\center
\begin{minipage}{5cm}
\center
\includegraphics[scale=0.3]{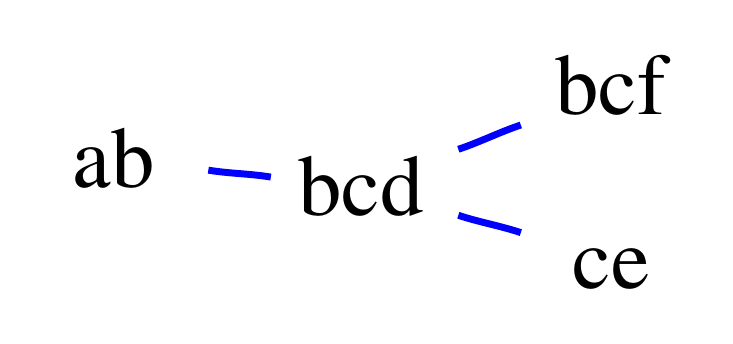}
\end{minipage}
\begin{minipage}{5cm}
\[
p(abcdef) = \frac{p(ab)p(bcd)p(bcf)p(ce)}{p(b)p(bc)p(c)}
\]
\end{minipage}
\label{fig:intro}
\caption{An example of a junction tree and the corresponding distribution decomposition.}
\end{figure}

As a simple example, consider a dataset $D$ with six attributes
$a,\ldots,f$, and a family $\cover{G}$ = \{$bcd$, $bcf$, $ab$, $ce$,
$bc$, $bd$, $cd$, $bf$, $cf$, $a$, $b$, $c$, $d$, $e$, $f$\}. The
family $\cover{G}$ can be represented as the junction tree shown in
Figure~\ref{fig:intro} such that the nodes in the tree are the maximal
itemsets in $\cover{G}$. Furthermore, the junction tree defines a
decomposable model of the dataset $D$.
 
Using decomposable itemset families has several notable
advantages. First of all, the following junction tree graphs provide
an extremely intuitive representation of the mining results. This is a
significant advantage over many other itemset selection methods, as
even small mining results of, say 50 itemsets, can be hard for humans
to grasp as a whole, if just plainly enumerated. Second, from the
computational point of view, decomposable families of itemsets provide
leverage for accurately solving problems that could be intractable
using the entire result set. Such problems include, for instance,
querying for frequency bounds of arbitrary attribute
combinations. Third, the statistical nature of the overall model
enable to incorporated regularization terms, like BIC, AIC, or MDL to
select only itemsets that reflect true dependencies between
attributes.

In this study we provide an efficient algorithm to build decomposable
itemset families while optimizing the likelihood of the model. We also
demonstrate how to use decomposable itemset families to execute
frequency bound querying, an intractable problem in the general
case. We provide empirical results showing that our algorithm works
well in practice.

The rest of the paper is organized as follows. Preliminaries are given
in Section~\ref{sec:prel} and the concept of decomposable models are
defined in Section~\ref{sec:model}. A greedy search algorithm is given
in Section~\ref{sec:algorithm}. Section~\ref{sec:experiments} is
devoted to experiments.  We present the related work in
Section~\ref{sec:related} and conclude the paper with discussion in
Section~\ref{sec:conclusions}. The proofs for the theorems in this
paper are provided in \cite{techraport}.

\section{Preliminaries and Notations}
\label{sec:prel}
In this section we describe the notation and the background definitions that are
used in the subsequent sections.

A \emph{binary dataset} $D$ is a collection of $N$ \emph{transactions}, binary
vectors of length $K$.  The dataset can be viewed as a binary matrix of size
$N \times K$. We denote the number of transactions by $\abs{D} = N$. The $i$th
element of a random transaction is represented by an \emph{attribute} $a_i$, a
Bernoulli random variable. We denote the collection of all the attributes by $A
= \enset{a_1}{a_K}$. An \emph{itemset} $X = \enset{x_1}{x_L} \subseteq A$ is a
subset of attributes. We will often use the dense notation $X = \iset{x_1}{x_L}$.

Given an itemset $X$ and a binary
vector $v$ of length $L$, we use the notation $p\pr{X = v}$ to express the
probability of $p\pr{x_1 = v_1, \ldots, x_L = v_L}$. If $v$ contains only 1s,
then we will use the notation $p\pr{X = 1}$.

Given a binary dataset $D$ we define $q_D$ to be an \emph{empirical distribution},
\[
q_D\pr{A = v} = \abs{\set{t \in D; t = v}} / \abs{D}.
\]
We define the frequency of an itemset to be $\freq{X} = q_D\pr{X = 1}$.
The \emph{entropy} of an itemset $X = \iset{x_1}{x_L}$ given $D$,
denoted by $\ent{X; D}$, is defined as
\begin{equation}
\ent{X; D} = -\sum_{v \in \set{0, 1}^L} q_D\pr{X = v} \log q_D\pr{X = v},
\label{eq:entropy}
\end{equation}
where the usual convention $0 \log 0 = 0$ is used. We omit $D$, when
it is clear from the context.

We say that a family $\cover{F}$ of itemsets is \emph{downward closed} if
each subset of a member of $\cover{F}$ is also included in $\cover{F}$.
An itemset $X \in \cover{F}$ is maximal if there is no $Y \in \cover{F}$ such
that $X \subset Y$.

\section{Decomposable Families of Itemsets}
\label{sec:model}
In this section we define the concept of decomposable families. Itemsets of a
decomposable family form a junction tree, a concept from the theory of Markov
Random Fields~\cite{cowell99network}. 

Let $\cover{G} = \enset{G_1}{G_M}$ be a downward closed family of itemsets covering the
attributes $A$. Let $H$ be a graph containing $M$ nodes where the $i$th node
corresponds to the itemset $G_i$. Nodes $G_i$ and $G_j$ are connected if $G_i$
and $G_j$ have a common attribute. The graph $H$ is called the \emph{clique graph}
and the nodes of $H$ are called \emph{cliques}.

We are interested in spanning trees of $H$ having a \emph{running intersection
property}. To define this property let $\tree{T}$ be a spanning tree of $H$.
Let $G_i$ and $G_j$ be two sets having a common attribute, say, $a$.  These
sets are connected in $\tree{T}$ by a unique path. Assume that $a$ occurs
in every $G_k$ along the path from $G_i$ to $G_j$. If this holds for
any $G_i$, $G_j$, and any common attribute $a$, then we say that the tree has a
running intersection property. Such a tree is called a \emph{junction tree}.

We should point out that the clique graph can have multiple junction trees and
that not all spanning trees are junction trees. In fact, it may be the case
that the clique graph does not have junction trees at all. If, however, the
clique graph has a junction tree, we call the original family $\cover{G}$
\emph{decomposable}.

We label edge $\pr{G_i, G_j}$ of a given junction tree
$\tree{T}$ with a set of mutual attributes $G_i \cap G_j$. This label set is
called a \emph{separator}.  We denote the set of all separators by
$\sep{\tree{T}}$. Furthermore, we denote the cliques of the tree by
$\vrt{\tree{T}}$.

Given a junction tree $\tree{T}$ and a binary vector $v$, we define the probability of $A = v$
to be 
\begin{equation}
\label{eq:junctiondistr}
p\pr{A = v; \tree{T}} =  \prod_{X \in \vrt{\tree{T}}} q_D\pr{X = v_{X}} \Big/ \prod_{Y \in \sep{\tree{T}}} q_D\pr{Y = v_{Y}}.
\end{equation}
It is a known fact that the distribution given in
Eq.~\ref{eq:junctiondistr} is actually the unique maximum entropy
distribution~\cite{jirousek95iterative,csiszar75divergence}.  Note
that $p\pr{A = v; \tree{T}}$ can be computed from the frequencies of
the itemsets in $\cover{G}$ using the inclusion-exclusion principle.

It can be shown that the family $\cover{G}$ is decomposable if and
only if the maximal sets of $\cover{G}$ is decomposable and that
Eq.~\ref{eq:junctiondistr} for the maximal sets of $\cover{G}$ and the
whole $\cover{G}$.  Hence, we usually construct the tree using only
the maximal sets of $\cover{G}$\footnote{We keep the family
$\cover{G}$ always downward closed.}. However, in some cases it is
convenient to have non-maximal sets as the cliques.  We will refer to
such cliques as \emph{redundant}.

Calculating the entropy of the tree $\tree{T}$ directly from
Eq.~\ref{eq:junctiondistr} gives us
\[
\ent{\tree{T}} = \sum_{X \in \vrt{\tree{T}}} \ent{X} - \sum_{Y \in \sep{\tree{T}}} \ent{Y}.
\]

A direct calculation from Eqs.~\ref{eq:entropy}--\ref{eq:junctiondistr} reveals that
$\log p\pr{D; \tree{T}} = -\abs{D} \ent{\tree{T}}$.
Hence, maximizing the log-likelihood of the data given $\tree{T}$ (whose components
are derived from the same data), is equivalent to minimizing the entropy.

We can define the maximum entropy distribution for any cover $\cover{F}$ via linear
constraints~\cite{csiszar75divergence}. The downside of this general approach
is that solving such a distribution is a \pp-hard
problem~\cite{tatti06complexity}.

The following definition will prove itself useful in subsequent sections.
Given two downward closed covers $\cover{G}_1$ and $\cover{G}_2$. We say that
$\cover{G}_1$ \emph{refines} $\cover{G}_2$, if $\cover{G}_1 \subseteq \cover{G}_2$.

\begin{proposition}
\label{thr:refinement}
If $\cover{G}_1$ refines $\cover{G}_2$, then $\ent{\cover{G}_1} \geq \ent{\cover{G}_2}$.
\end{proposition}

\section{Finding Trees with Low Entropy}
\label{sec:algorithm}
In this section we describe the algorithm for searching decomposable families.
To be more precise, given a candidate set, a downward closed family $\cover{F}$
covering the set of attributes $A$, our goal is to find a decomposable downward
closed family $\cover{G} \subseteq \cover{F}$. Hence our goal is to find a
junction tree $\tree{T}$ whose cliques are $\cover{G}$.

\subsection{Definition of the Algorithm}

We search the tree in an iterative fashion. At the beginning of each iteration
round we have a junction tree $\tree{T}_{n - 1}$ whose cliques have at most $n$ attributes.
The first tree is $\tree{T}_0$ containing only single attributes and no edges.
During each round the tree is modified so that in the end we will have
$\tree{T}_n$, a tree with cliques having at most $n + 1$ attributes.

In order to fully describe the algorithm
we need the following definition: $X$ and $Y$ are said to be \emph{$n-1$-connected} 
in a junction tree $\tree{T}$, if there is a path in $\tree{T}$ from $X$ to $Y$
having at least one separator of size $n-1$. We say that $X$ and $Y$ are $0$-connected,
if $X$ and $Y$ are not connected.

Each round of the algorithm consists of three steps. The pseudo-code of the algorithm
is given in Algorithm~\ref{alg:searchtree}--\ref{alg:modifytree}.
\begin{enumerate}
\item \textbf{Generate:} We construct a graph $G_n$ whose nodes are the cliques
of size $n$ in $\tree{T}_{n - 1}$. We add all the edges to $G_n$ having the form $\pr{X, Y}$ 
such that $\abs{X \cap Y} = n - 1$ and $X \cup Y \in \cover{F}$.
We also set $\tree{T}_n = \tree{T}_{n - 1}$. The weight of the edge is
set to
\[
w\pr{e} = \ent{X} + \ent{Y} - \ent{X \cap Y} - \ent{X \cup Y}.
\]
\item \textbf{Augment:} We select the edge, say $e = \pr{X, Y}$, having
the largest weight and remove it from $G_n$. If $X$ and
$Y$ are $n - 1$ -connected in $\tree{T}_n$ we add
$\tree{T}_n$ with a new clique $V = X \cup Y$. Furthermore, for each
$v \in V$, we consider $W = V - v$. If $W$ is not in $\tree{T}_n$, it
is added into $\tree{T}_n$. Next, $W$ and $V$ are connected in
$\tree{T}_n$. At the same time, the
node $W$ is also added into $G_n$ and the edges of $G_n$ are added
using the same criteria as in Step 1 (Generate).
Finally, a possible cycle is removed from $\tree{T}_n$ by
finding an edge with separator of size $n - 1$.
Augmenting is
repeated as long as $G_n$ has no edges.

\item \textbf{Purge:} The set $\vrt{\tree{T}_n}$ contains redudant cliques after augmentation.
We purge the tree by removing the redudant cliques of $\tree{T}_n$.
\end{enumerate}

To illustrate the algorithm we provide a toy example.

\begin{example}
Consider that we have a family 
\[
\cover{F} = \set{a, b, c, d, e, ab, ac, ad, bc, bd, cd, ce, abc, acd, bcd}.
\]

Assume that we are at the beginning of the second round and we already have the
junction tree $\tree{T}_1$ given in Figure~\ref{fig:searchtree_example:a}. We
form $G_2$ by taking the edges $\pr{ab, bc}$ and $\pr{bc, cd}$.

Consider that we pick $ab$ and $bc$ for joining. This will spawn $ac$
and $abc$ in $\tree{T}_2$ (Figure~\ref{fig:searchtree_example:d}) and
$ac$ in $G_2$ (Figure~\ref{fig:searchtree_example:e}). Note that we
also add the edge $\pr{ac, cd}$ into $G_2$.  Assume that we continue
by picking $\pr{ac, cd}$. This will spawn $acd$ and $cd$ into
$\tree{T}_2$. Note that $\pr{bc, cd}$ is removed from $\tree{T}_2$
in order to break the cycle.

The last edge $\pr{bc, cd}$ is not added into $\tree{T}_2$ since $bc$
and $cd$ are not $n-1$-connected.  The final tree
(Figure~\ref{fig:searchtree_example:g}) is obtained by keeping only
the maximal sets, that is, purging the cliques $bc$, $ab$, $ac$,
$ad$, and $cd$. The corresponding decomposable family is $\cover{G} = \cover{F} - bcd$.

\begin{figure}[htb!]
\center
\subfigure[$\tree{T}_2$ at the beginning.]
	{\label{fig:searchtree_example:a}\begin{minipage}[b]{4cm}\center\includegraphics[scale=0.3]{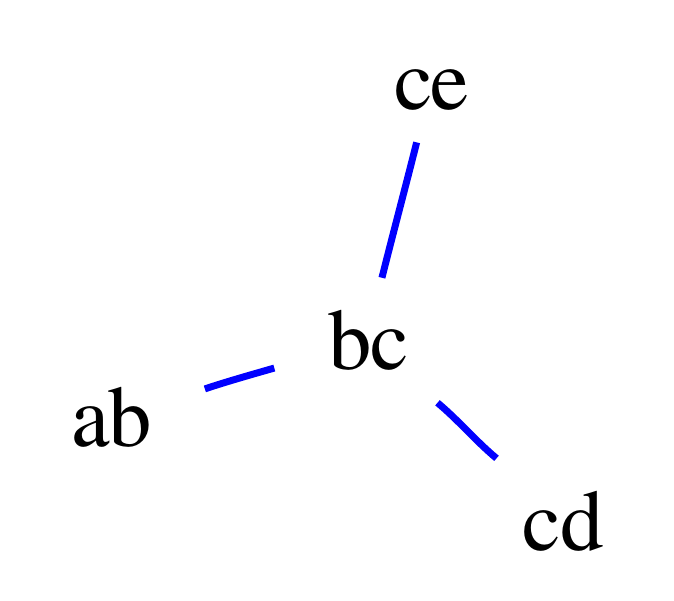}\end{minipage}}%
\subfigure[$G_2$ at the beginning.]
	{\label{fig:searchtree_example:c}\begin{minipage}[b]{4cm}\center\includegraphics[scale=0.3]{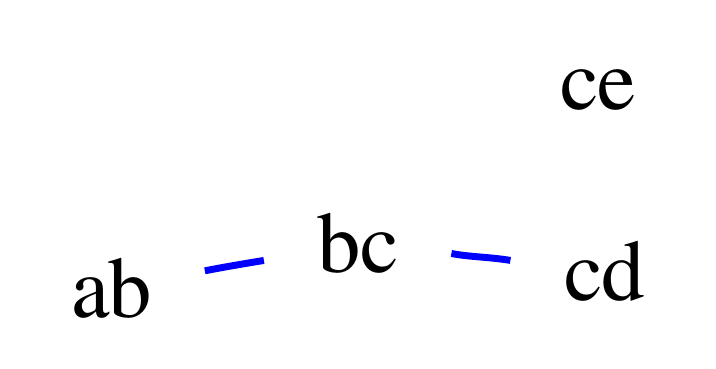}\end{minipage}}%
\subfigure[$\tree{T}_2$, $ab$ and $bc$ joined.]
	{\label{fig:searchtree_example:d}\begin{minipage}[b]{4cm}\center\includegraphics[scale=0.3]{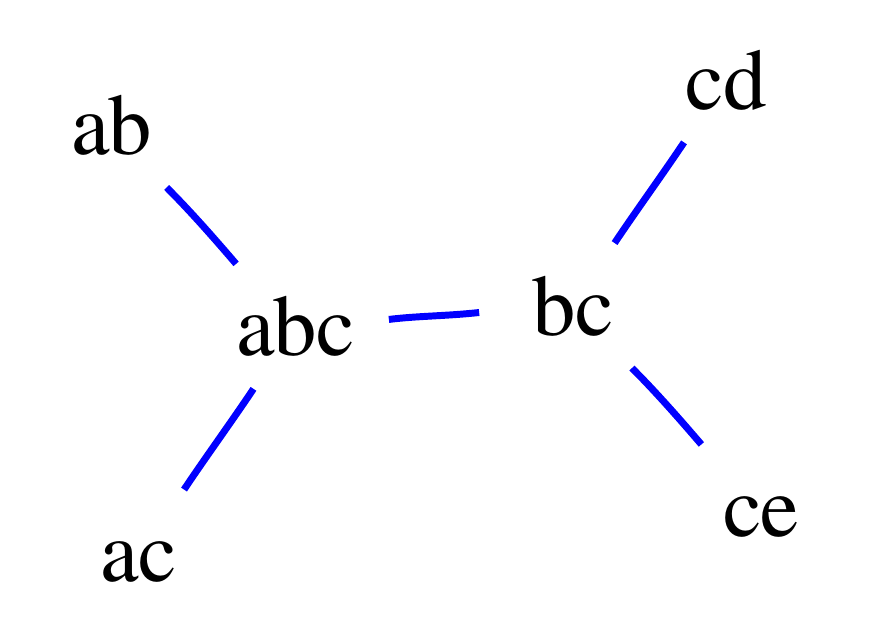}\end{minipage}}
\subfigure[$G_2$ $ab$ and $bc$ joined.]
	{\label{fig:searchtree_example:e}\begin{minipage}[b]{4cm}\center\includegraphics[scale=0.3]{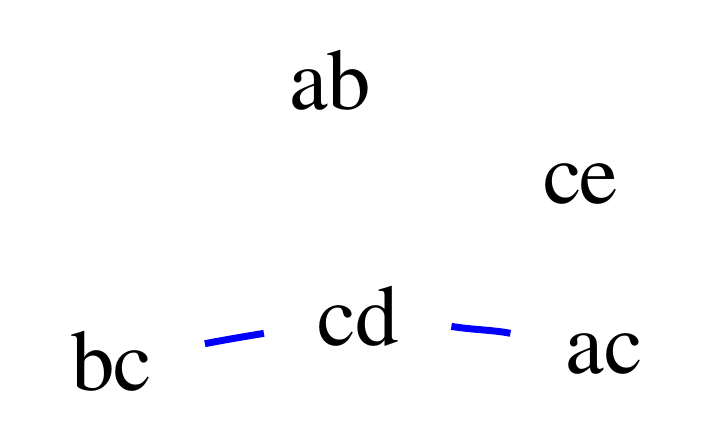}\end{minipage}}%
\subfigure[$\tree{T}_2$ after joining $ac$ and $cd$.]
	{\label{fig:searchtree_example:f}\includegraphics[scale=0.3]{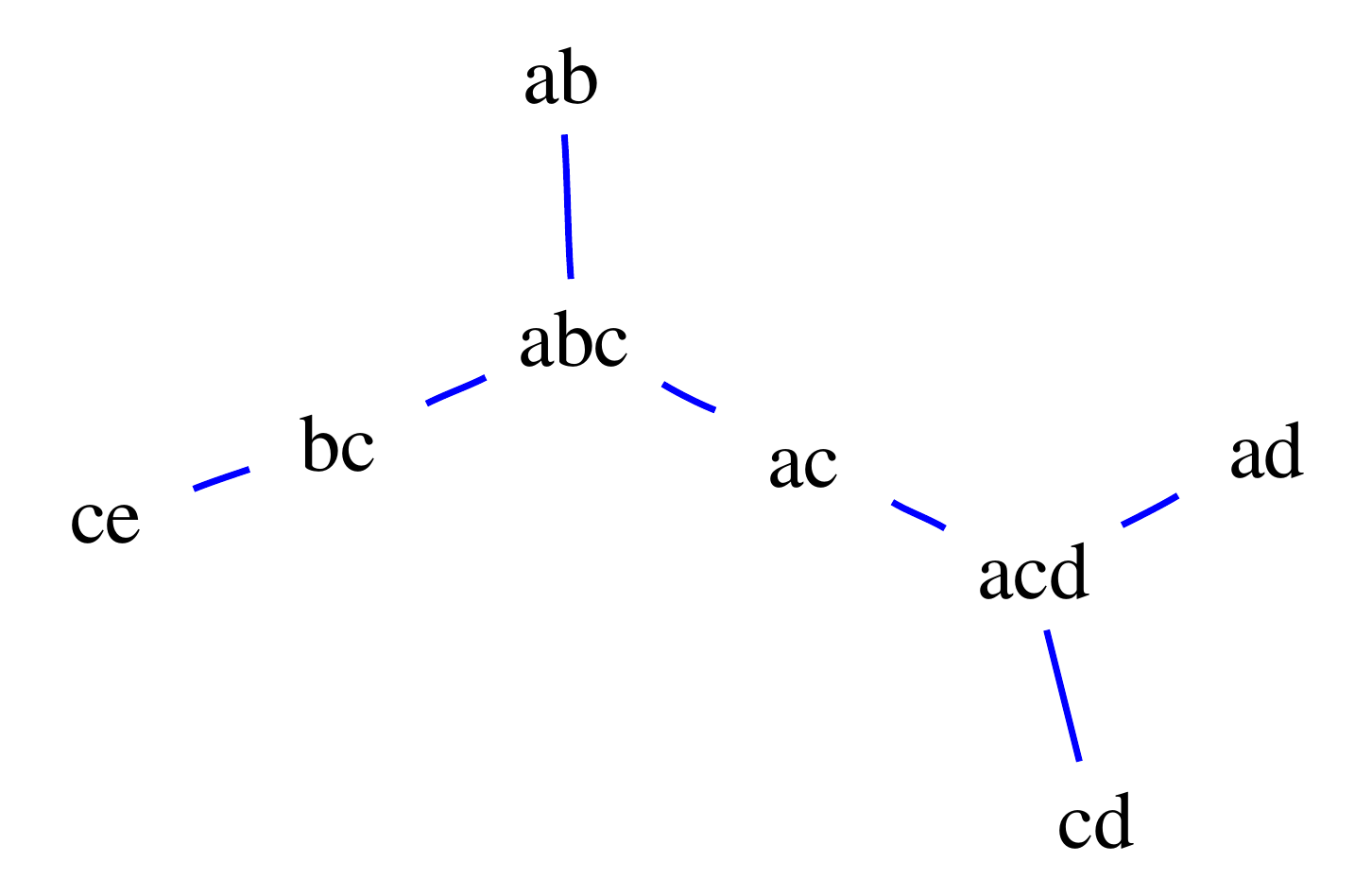}}%
\subfigure[Final $\tree{T}_2$]
	{\label{fig:searchtree_example:g}\includegraphics[scale=0.3]{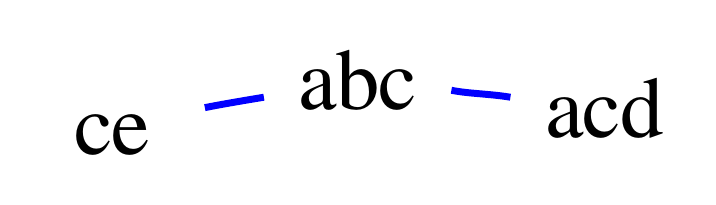}}
\caption{Example of graphs during different stages of \textsc{SearchTree} algorithm.}
\label{fig:searchtree_example}
\end{figure}

\end{example}

\begin{algorithm}[ht!]
\caption{\textsc{SearchTree} algorithm. The input is a downward closed cover
$\cover{F}$, the output is a junction tree $\tree{T}$ such that $\vrt{\tree{T}} \subseteq
\cover{F}$.}
\begin{algorithmic}
\STATE $\vrt{\tree{T}_0} \define \set{x; x \in A}$ \COMMENT{$\tree{T}_0$ contains the single items.}
\STATE $n \define 0$.

\REPEAT
\STATE $n \define n + 1$.
\STATE $\tree{T}_n \define \tree{T}_{n - 1}$.
\STATE $\vrt{G_n} \define \set{X \in \vrt{\tree{T}_n}; \abs{X} = n}$.
\STATE $\edge{G_n} \define \set{\pr{X, Y}; X, Y \in \vrt{G_n}, \abs{X \cap Y} = n - 1, X \cup Y \in \cover{F}}$.

\REPEAT
\STATE $e = \pr{X, Y} \define \arg\max_{x \in \edge{G_n}} w(x)$. 
\STATE $\edge{G_n} \define \edge{G_n} - e$.

\IF{$X$ and $Y$ are $n-1$-connected in $\tree{T}_n$}
\STATE Call \textsc{ModifyTree}.
\ENDIF

\UNTIL{$\edge{G_n} = \emptyset$}
\STATE Delete marked nodes from $\tree{T}_n$, connect the incident nodes.
\UNTIL{$G_n$ is empty}
\STATE \textbf{return} $\tree{T}$
\end{algorithmic}
\label{alg:searchtree}
\end{algorithm}

\begin{algorithm}[ht!]
\caption{\textsc{ModifyTree} algorithm.}
\begin{algorithmic}
\STATE $V \define X \cup Y$.
\STATE $\vrt{\tree{T}_n} \define \vrt{\tree{T}_n} + V$.
\FOR{$v \in V$}
\STATE $W \define V - v$.
\STATE Mark $W$.
\IF{$W \notin \vrt{G_n}$}

\STATE $\vrt{G_n} \define \vrt{G_n} + W$.
\STATE $\edge{G_n} \define \edge{G_n} + \set{\pr{W, Z}; Z \in \vrt{G_n}, \abs{X \cap Z} = n - 1, V \neq X \cup Z \in \cover{F}}$.
\STATE $\vrt{\tree{T}_n} \define \vrt{\tree{T}_n} + W$.

\ENDIF
\STATE $\edge{\tree{T}_n} \define \edge{\tree{T}_n} + \pr{V, W}$.
\ENDFOR
\STATE Remove the possible cycle in $\tree{T}_n$ by removing an edge $\pr{U, V}$ connecting
$X$ and $Y$ and having $\abs{U \cap V} = n - 1$.

\end{algorithmic}
\label{alg:modifytree}
\end{algorithm}

The next theorem states that the \textbf{Augment} step does not violate the running intersection
property.

\begin{theorem}
Let $\tree{T}$ be a junction tree with cliques of size $n + 1$, at maximum.
Let $X, Y \in \vrt{T}$ be cliques of size $n$ such that $\abs{X \cap Y} = n - 1$.
Set $V = X \cup Y$. Then the family $\vrt{\tree{T}} + V$ is decomposable if and only
if $X$ and $Y$ are $n-1$-connected in $\tree{T}$.
\label{thr:junctionlegal}
\end{theorem}
\begin{theorem}
\textsc{ModifyTree} decreases the entropy of $\tree{T}_n$ by $w(e)$.
\label{thr:edgeweight}
\end{theorem}

Theorems~\ref{thr:junctionlegal}--\ref{thr:edgeweight} imply that \textsc{SearchTree}
algorithm is nothing more than a greedy search. However, since we are adding cliques
in rounds we can state that under some conditions the algorithm returns an optimal cover
for each round.

\begin{theorem}
\label{thr:coveroptimal}
If all the members of $\cover{F}$ of size $n + 1$ are added to $G_n$
at the beginning of the $n$th round, then $\tree{T}_{n+1}$ has the
lowest entropy among the families refined by $\tree{T}_n$ and containing
the sets of size $n + 1$, at maximum.
\end{theorem}

\begin{corollary}
\label{cor:chowliu}
The tree $\cover{T}_1$ is optimal among the families using the sets of size $2$.
\end{corollary}

Corollary~\ref{cor:chowliu} states that $\cover{G}_1$ is the Chow-Liu tree~\cite{chow68treemodel}.

\subsection{Model Selection}
Theorem~\ref{thr:refinement} reveals a drawback in the current approach. Consider
that we have two independent items $a$ and $b$ and that $\cover{F} = \set{a, b, ab}$. 
Note that $\cover{F}$ is itself decomposable and $\cover{G} = \cover{F}$. However,
a more reasonable family would be $\set{a, b}$. To remedy this problem we will use model
selection techniques such as BIC~\cite{schwarz78bic}, AIC~\cite{akaike74aic}, and
Refined MDL~\cite{grunwald07mdl}. All these methods score the model by adding a penalty
term to the likelihood.

We modify the algorithm by considering only the edges in $G_n$ that improve the score.
For BIC this reduces to considering only the edges satisfying
\[
\abs{D}w(e) \geq 2^{n - 2}\log \abs{D},
\]
where $n$ is the current level of \textsc{SearchTree} algorithm. Using AIC leads to
the considering only the edges for which
\[
\abs{D}w(e) \geq 2^{n - 1}.
\]

Refined MDL is more troublesome. The penalty term in MDL is known as
\emph{stochastic complexity}. In general, there is no known closed
formula for the stochastic complexity, but it can be computed for the
multinomial distribution in linear
time~\cite{kontkanen07linear}. However, it is numerically unstable for
data with large number of transactions. Hence, we will apply
often-used asympotic estimate~\cite{rissanen96fisher} and define the penalty term
\[
\costmdl{k} = \frac{k - 1}{2}\log \abs{D} - \frac{1}{2} \log \pi - \log \mathrm{\Gamma}\pr{k /2}
\]
for $k$-multinomial distribution.

There are no known exact or approximative solution in a closed form of stochastic complexity for
junction trees. Hence we propose the penalty term for the tree to be
\[
\sum_{X \in \vrt{\tree{T}}} \costmdl{2^{\abs{X}}} - \sum_{Y \in \sep{\tree{T}}} \costmdl{2^{\abs{Y}}}.
\]
Here we think that a single clique $X$ is a $2^{\abs{X}}$-multinomial distribution and
we compensate the possible overlaps of the cliques by subtracting the complexity of
the separators. Using this estimate leads to a selection criteria
\[
\abs{D}w(e) \geq \costmdl{2^{\abs{n + 1}}} -  2\costmdl{2^{\abs{n}}} + \costmdl{2^{\abs{n - 1}}}.
\]

\subsection{Computing Multiple Decomposable Families}
\label{sec:multiple}
We can use \textsc{SearchTree} algorithm for computing multiple decomposable
covers from a single candidate set $\cover{F}$. The motivation behind this approach is
that we are able to use more itemsets. We will show empirically in Section~\ref{sec:experiments:query}
that the bounds for boolean queries (see Section~\ref{sec:query}) improve
significantly if we are using multiple covers.

Set $\cover{F}_1 = \cover{F}$ and let $\cover{G}_1$ be the
first decomposable family constructed from $\cover{F}_1$ using \textsc{SearchTree}
algorithm. We define
\[
\cover{F}_2 = \cover{F}_1 - \set{X \in \cover{F}_1; \text{ there is } Y \in \cover{G}_1, \abs{Y} > 1, Y \subseteq X}.
\]
We compute $\cover{G}_2$ from $\cover{F}_2$ and continue in the iterative fashion
until $\cover{G}_k$ contains nothing but individual items. 

\section{Boolean Queries with Decomposable Families}
\label{sec:query}
One of our motivations for constructing decomposable families is that
some computational problems that are hard for general families of
itemsets reduce to tractable if the underlying family is
decomposable. In this section we will show that the computational
burden of a boolean query, a classic optimization
problem~\cite{hailperin65inequalities,bykowski04support}, reduces
significantly, if we are using decomposable families of itemsets.

Assume that we are given a set of known itemsets $\cover{G}$ and a
query itemset $Q \notin \cover{G}$. We wish to find $\freq{Q;
\cover{G}}$, the possible frequencies for $Q$ given the frequencies of
$\cover{G}$.  It is easy to see that the frequencies form an interval,
hence it is sufficient to find the maximal and the minimal
frequencies. We can express the problem of finding the maximal
frequency as a search for the distribution $p$ solving
\begin{equation}
\label{eq:maxfreq}
\begin{array}{rl}
\max & p\pr{Q = 1} \\
\text{s.t.} & p\pr{X = 1} = \freq{X}, \text{ for each } X \in \cover{G}. \\
& p \text{ is a distribution over } A.
\end{array}
\end{equation}

We can solve Eq.~\ref{eq:maxfreq} using Linear
Programming~\cite{hailperin65inequalities}. However, the number of
variables in the program is $2^{\abs{A}}$ and makes the program tractable
only for small datasets. In fact, solving Eq.~\ref{eq:maxfreq} is an
\np-hard problem~\cite{tatti06complexity}.

In the rest of the section we present a method of solving
Eq.~\ref{eq:maxfreq} with a linear program containing only
$2^{\abs{Q}}\abs{\cover{G}}\abs{A}$ variables, assuming that
$\cover{G}$ is decomposable.  This method is an explicit construction
of the technique presented in~\cite{tatti06projections}.  The idea
behind the approach is that instead of solving a joint distribution in
Eq.~\ref{eq:maxfreq}, we break the distribution into small component
distributions, one for each clique in the junction tree.  These
components are forced to be consistent by requiring that they are
equal at the separators.  The details are given in
Algorithm~\ref{alg:query}.

\begin{algorithm}
\begin{algorithmic}
\STATE $\enset{\tree{T}_1}{\tree{T}_M} \define$ connected components of a junction tree of $\cover{G}$.
\FOR{$i = 1,\ldots,M$}
\STATE $Q_i \define Q \cap \pr{\bigcup \vrt{\tree{T}_i}}$. \COMMENT{Items of $Q$ contained in $\tree{T}_i$.}
\STATE $\tree{U} \define \arg \min_{\tree{S} \subseteq \tree{T}_i} \set{\abs{\vrt{\tree{S}}} ; Q_i \subseteq \bigcup \vrt{\tree{S}}}$.
\COMMENT{Smallest subtree containing $Q_i$.}
\WHILE{there are changes}
\STATE Remove the items outside $Q_i$ that occur in only one clique of $\tree{U}$.
\STATE Remove redundant cliques.
\ENDWHILE
\STATE Select one clique, say $R$ from $\tree{U}$ to be the root.
\STATE $R \define R \cup Q_i$. \COMMENT{Augment the root with $Q_i$}
\STATE Augment the rest cliques in $\tree{U}$ so that the running intersection property holds.
\STATE Let $p_C$ be a distribution over each clique $C \in \vrt{\tree{U}}$.
\STATE $\alpha_i \define $ the solution of a linear program
\[
\begin{array}{rl}
\min & p_R\pr{Q_i = 1} \\
\text{s.t.} & p_C\pr{X = 1} = \freq{X}, \text{ for each } C \in \vrt{\tree{U}}, X \in \cover{G}, X \subseteq C. \\
& p_{C_1}\pr{C_1 \cap C_2} = p_{C_2}\pr{C_1 \cap C_2}, \text{ for each } \pr{C_1, C_2} \in \edge{\tree{U}}. \\
\end{array}
\]
\STATE $\beta_i \define $ the solution of the maximum version of the linear program.
\ENDFOR
\STATE $\freq{Q; \cover{G}} \define \spr{\max\pr{\sum_i^M \alpha_i - (M - 1), \, 0}, \min_i\pr{\beta_i}}$.
\end{algorithmic}
\caption{\textsc{QueryTree} algorithm for solving a query $Q$ from a decomposable cover $\cover{G}$. The output is the interval $\freq{Q; \cover{G}}$.}
\label{alg:query}
\end{algorithm}

To clarify the process we provide the following simple example.

\begin{example}
Assume that we have $\cover{G}$ whose junction tree is given in Figure~\ref{fig:query:a}.
Let query be $Q = adg$. We begin first by finding the smallest sub-tree containing $Q$.
This results in purging $fh$ (Figure~\ref{fig:query:b}). We further purge the tree
by removing $e$ since it only occurs in one clique (Figure~\ref{fig:query:c}). In
the next step we pick a root, which in this case is $bc$ and augment the cliques
with the members of $Q$ so that the root contains $Q$ (Figure~\ref{fig:query:d}).
We finally remove the redundant cliques which are $ab$, $cd$, $fg$. The final
tree is given in~\ref{fig:query:e}. Finally, the linear program is formed using
two distributions $p_{abcdg}$ and $p_{cfg}$. The number of variables in this program
is $2^5 + 2^3 = 40$ opposed to the original $2^8 = 256$.

\begin{figure}[htb!]
\center
\subfigure[Original $\tree{T}$] 
    {\label{fig:query:a}\includegraphics[scale=0.3]{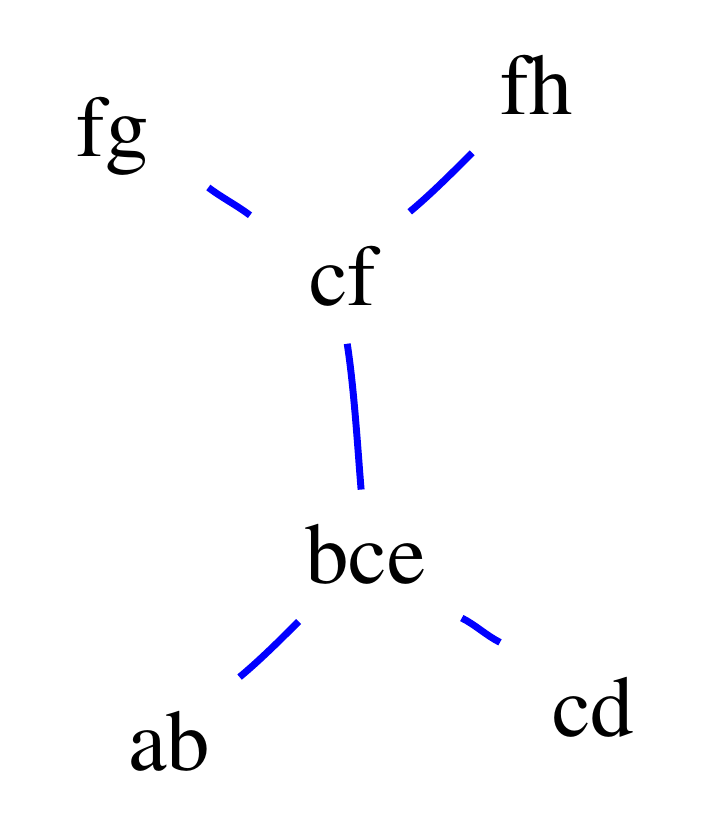}}%
\subfigure[$\tree{U}$] 
    {\label{fig:query:b}\includegraphics[scale=0.3]{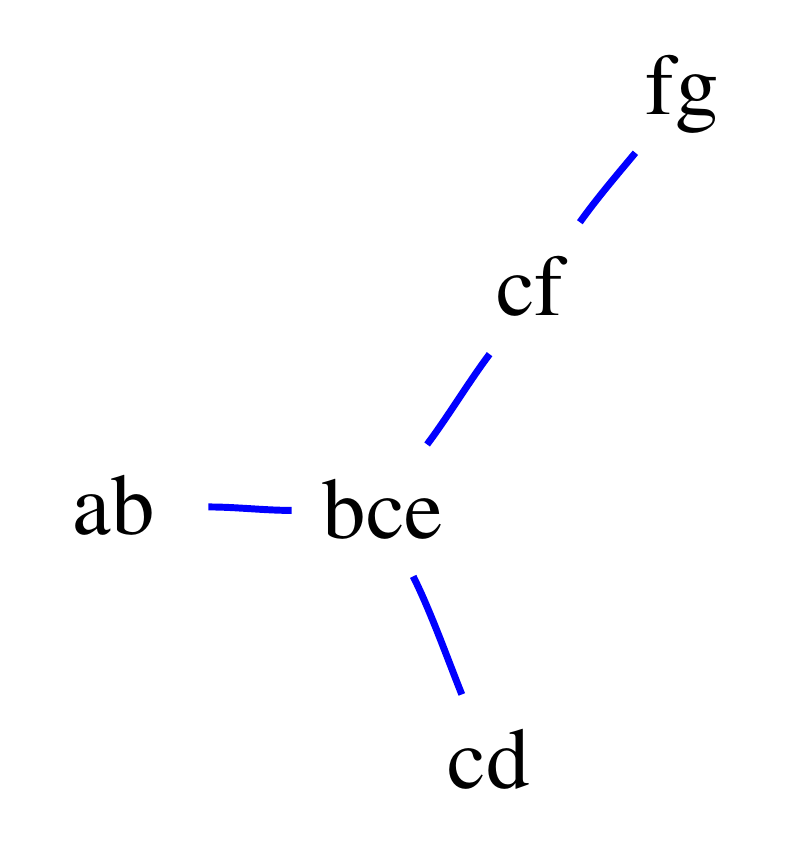}}%
\subfigure[Purged $\tree{U}$] 
    {\label{fig:query:c}\includegraphics[scale=0.3]{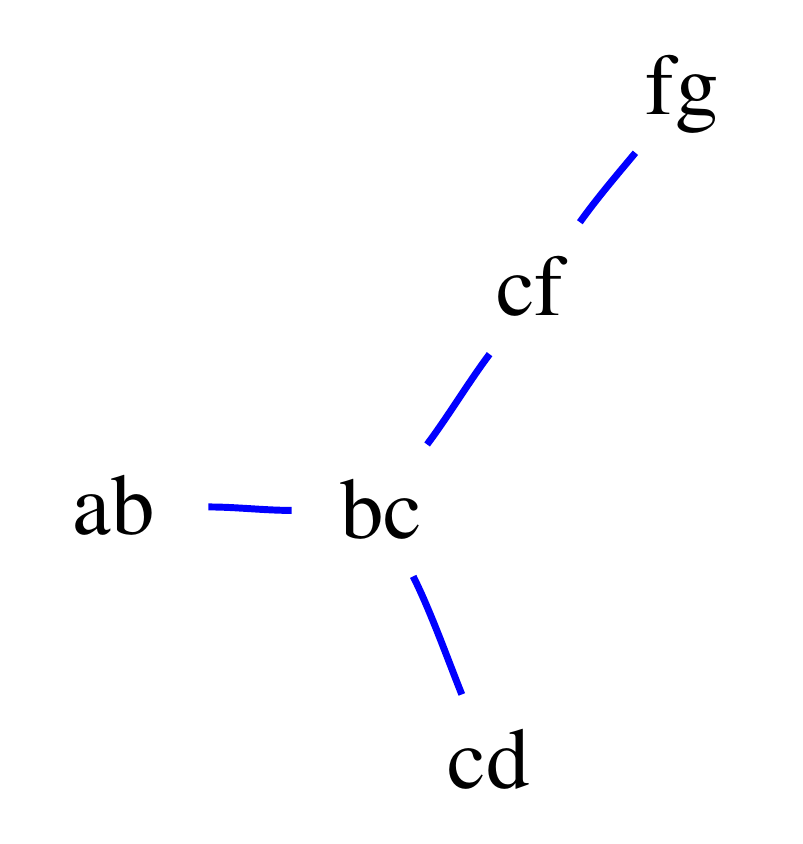}}%
\subfigure[Augmented $\tree{U}$] 
    {\label{fig:query:d}\begin{minipage}[b]{3cm}\center\includegraphics[scale=0.3]{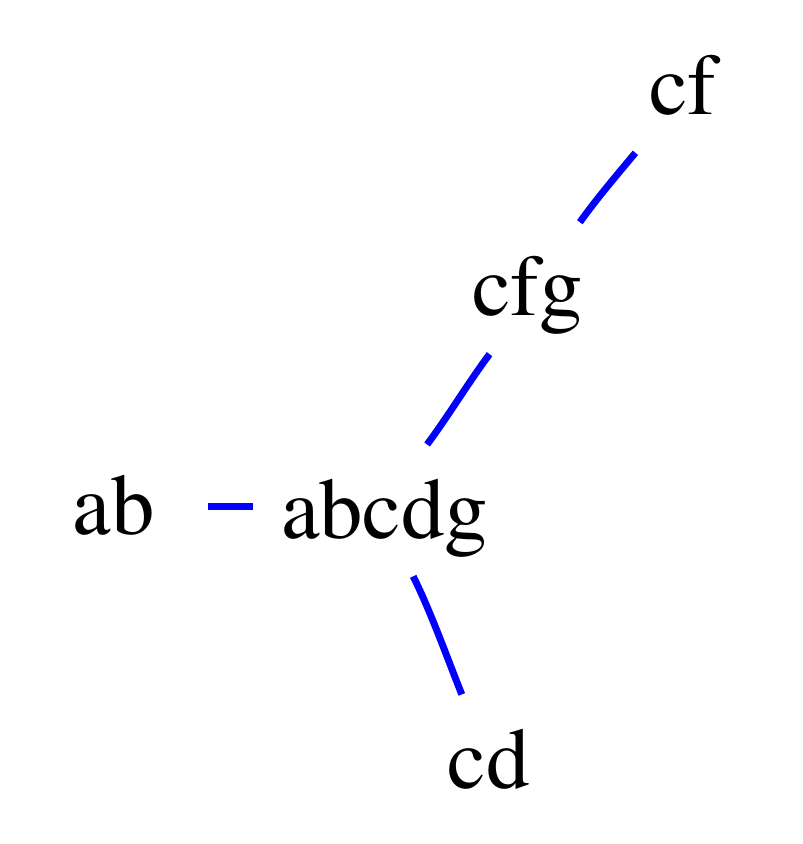}\end{minipage}}%
\subfigure[Final $\tree{U}$] 
    {\label{fig:query:e}\begin{minipage}[b]{2cm}\center\includegraphics[scale=0.3]{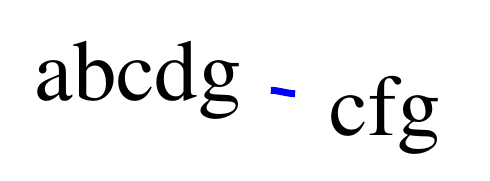}\end{minipage}}
\caption{Junction trees during different stages of solving the query problem.}
\end{figure}

\end{example}

Note that we did not specify in Algorithm~\ref{alg:query} which clique
we selected to be the root $R$. The linear program depends on the root $R$ and hence
we select the root minimizing the number of variables in the linear program.

\begin{theorem}
\label{thr:query}
\textsc{QueryTree} algorithm solves correctly the boolean query $\freq{q; \cover{G}}$.
The number of variables occurring in the linear programs is $2^{\abs{Q}}\abs{\cover{G}}\abs{A}$, at maximum.
\end{theorem}

\section{Experiments}
\label{sec:experiments}
In this section we will study empirically
the relationship between the decomposable itemset families and the candidate
set, the role of the regularization, and the performance of boolean
queries using multiple decomposable families.

\subsection{Datasets}
For our experiments we used one synthetic generated dataset,
\dtname{Path}, and three real-world datasets: \dtname{Paleo},
\dtname{Courses} and \dtname{Mammals}. The synthetic dataset,
\dtname{Path}, contained 8 items and 100 transactions. Each item was
generated from the previous item by flipping it with a $0.3$
probability. The first item was generated by a fair coin flip. The
dataset \dtname{Paleo}\footnote{NOW public release 030717 available
from~\cite{fortelius05now}.} contains information of mammal fossils
found in specific paleontological sites in
Europe~\cite{fortelius05now}. \dtname{Courses} describes computer
science courses taken by students at the Department of Computer
Science of the University of Helsinki. The
\dtname{Mammals}\footnote{The full version of the mammal dataset is
available for research purposes upon request from the Societas
Europaea Mammalogica (\texttt{www.european-mammals.org})} dataset
consists of presence/absence records of current day European mammals
\cite{mitchell-jones99}. The basic characteristics of the real-world
data sets are shown in Table~\ref{table:datasets}.

\begin{table}[htb!]
\begin{center}
\begin{tabular}{lccrr}
\toprule
Dataset & \# of rows &$\,\,\,$ \# of items  &$\,\,\,$ \# of 1s  & $\,$$\,$$\,$$\,$$\,$$\,$$\,$$ \frac{\text{\# of 1s}}{\text{\# of rows}}$ \\
\midrule
\dtname{Paleo} & 501 & 139 & 1980 & 16.0 \\ 
\dtname{Courses}  &  3506 & 98 & 16086 & 4.6 \\
\dtname{Mammals} & 2183 & 124 & 54155 & 24.8 \\
\bottomrule
\end{tabular}
\end{center}
\caption{\label{table:datasets}
The basic properties of the datasets.}
\end{table}

\subsection{Generating Decomposable Families}
\begin{figure}[htb!]
\center
\subfigure[First junction tree of \dtname{Path} data.]{\label{fig:path:a}\includegraphics[scale=0.3]{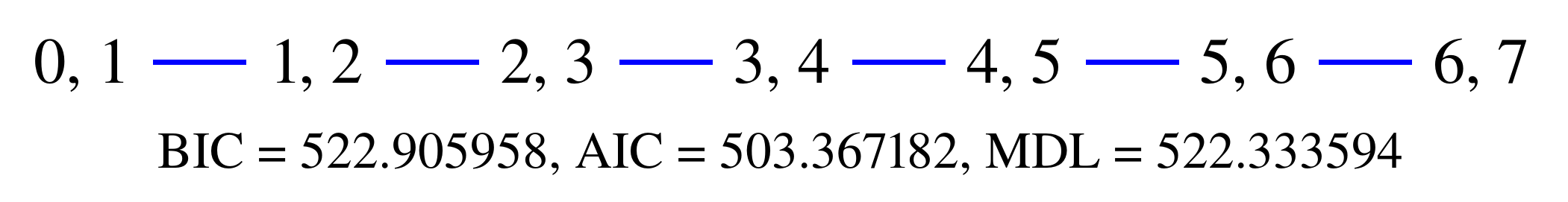}}
\subfigure[Second junction tree of \dtname{Path} data.]{\label{fig:path:b}\includegraphics[scale=0.3]{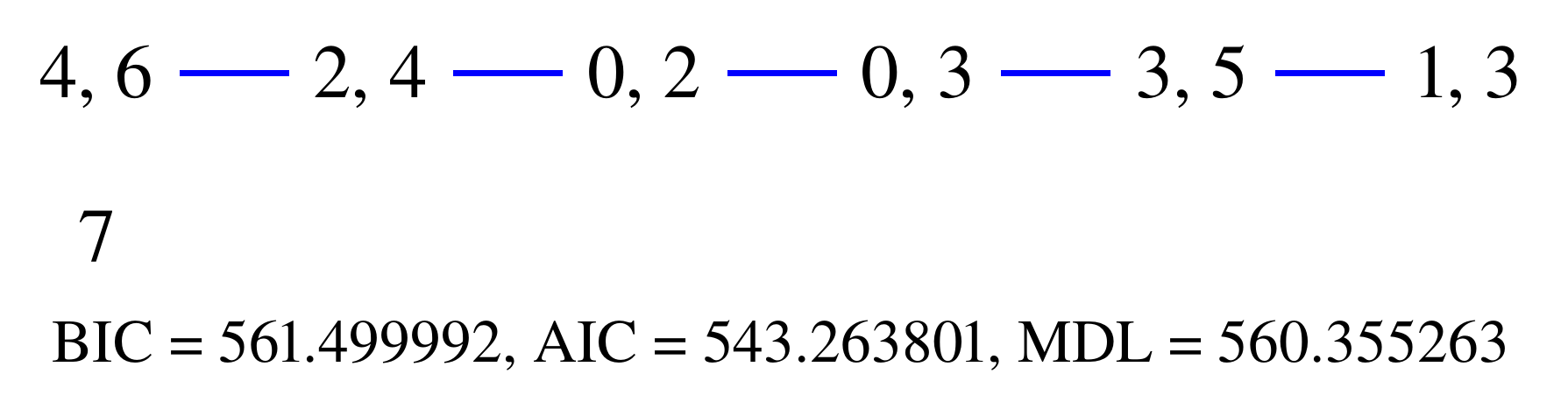}}
\caption{Junction trees for \dtname{Path}, a syntetic data in which an item is
generated from the previous item by flipping it with $0.3$ probability. The
junction trees are regularized using BIC. The tree in Figure~\ref{fig:path:b}
is generated by ignoring the cliques of the tree in Figure~\ref{fig:path:a}.}
\label{fig:path}
\end{figure}

In our first experiment we examined the junction trees that were
constructed for the \dtname{Path} dataset. We calculated a sequence of
trees using the technique described in Section~\ref{sec:multiple}. As
input to the algorithm we used an unconstrained candidate collection of
itemsets (minimum support = 0) from \dtname{Path} and BIC as the
regularization method. In Figure~\ref{fig:path:a} we see that the
first tree corresponds to the model used to generate the
dataset.  The second tree, given in Figure~\ref{fig:path:b}, tend to
link the items that are one gap away from each other. This is a
natural result since close items are the most informative about each
other.

With \dtname{Courses} data one large junction tree of itemsets is
produced with several noticeable components. One distinct component at
one end of the tree contains introductory courses like
\textit{Introduction to Programming}, \textit{Introduction to
Databases}, \textit{Introduction to Application design} and
\textit{Java Programming}. Respectively, the other end of the tree
features several distinct components with itemsets on more specialized
themes in computer science and software engineering. The central node
connecting each of these components in the entire tree is the itemset
node \{\textit{Software Engineering, Models of Programming and
Computing, Concurrent systems}\}.

\begin{figure}[htb!]
\center
\includegraphics[width=1\textwidth]{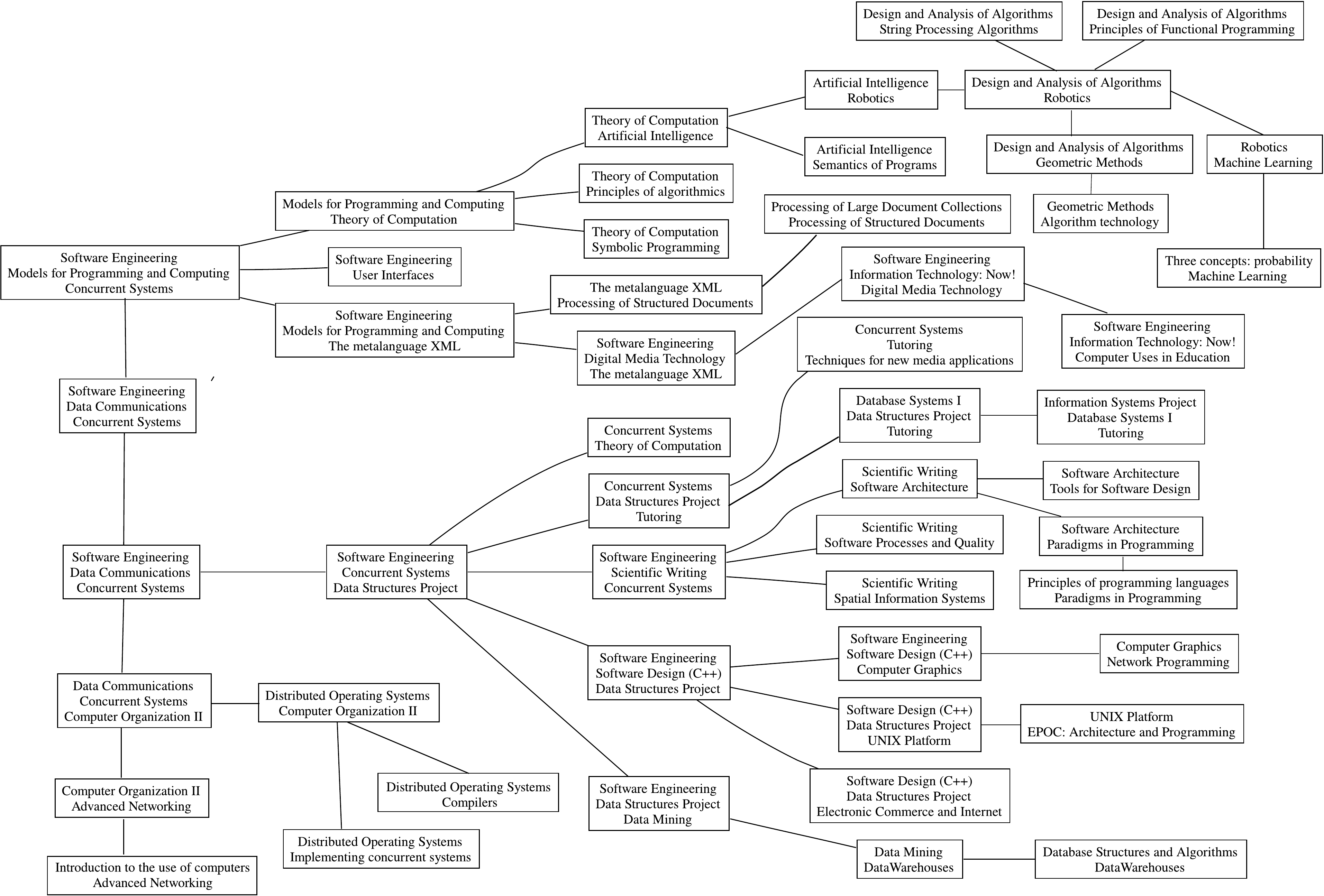}
\caption{A part of the junction tree constructed from the
\dtname{Courses} dataset. The tree was constructed using an
unconstrained candidate family (min. support = 0) as input and BIC as
regularization.}
\label{fig:example_course}
\end{figure}

Figure~\ref{fig:example_course} shows about two-thirds of the entire
\dtname{Courses} junction tree, with the component related to
introductory courses removed because of the space constraints.  We see
a concurrent and distributed systems related component in the lower
left part of the figure, a more software development oriented
component in the lower right quarter and a Robotics/AI component in
the upper right corner of the tree. The entire \dtname{Courses}
junction tree can be found from \cite{techraport}.

\begin{figure}[htb!]
\center
\subfigure[\dtname{Paleo}]{\includegraphics[width=4cm]{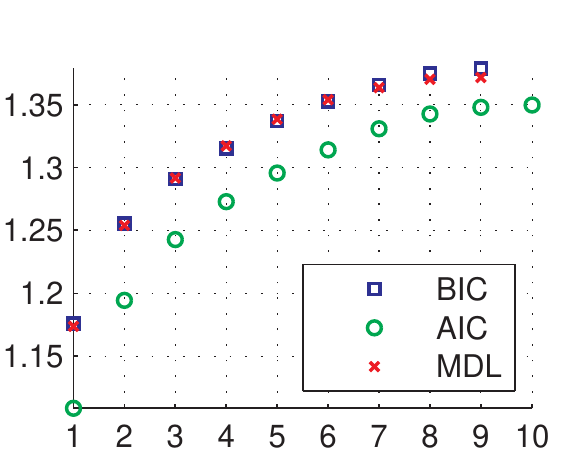}}
\subfigure[\dtname{Courses}]{\includegraphics[width=4cm]{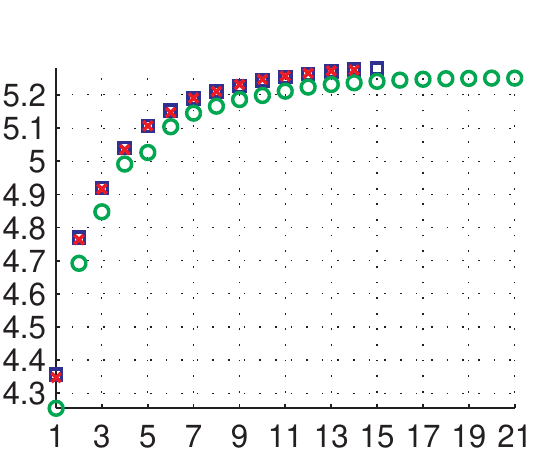}}
\subfigure[\dtname{Mammals}]{\includegraphics[width=4cm]{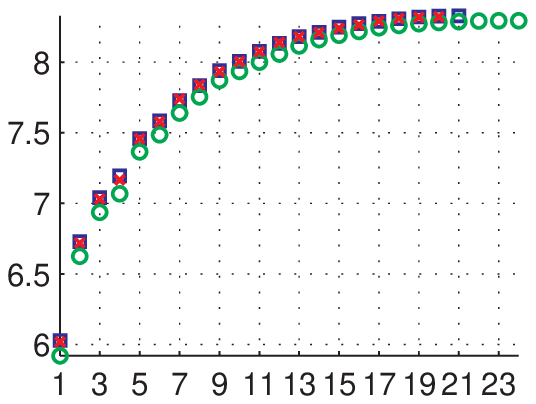}}

\caption{Scores of covers as a function of the order of the
cover. Each cover is computed with an unconstrained candidate family
(min. support = 0) as input and the corresponding regularization. The
$y$-axis is the model score divided by $10^4$.}
\label{fig:score}
\end{figure}

We continued our experiments by studying the behavior of the model
scores in a sequence of trees induced by a corresponding sequence of
decomposable families. For the \dtname{Path} data the scores of the
two first junction trees are shown in Figure~\ref{fig:path}, with the
first one yielding smaller values. For the real-world datasets, we
computed a sequence of trees from each dataset, again, with the
unconstrained candidate collection as input and using AIC, BIC, or MDL
respectively as the regularization method. Computation took about 1
minute per tree.  The corresponding scores are plotted as a function
of the order of the corresponding junction tree
(Figure~\ref{fig:score}). The scores are increasing in the sequence,
which is expected since the algorithm tries to select the best model
and the subsequent trees are constructed from the left-over
itemsets. The increase rate slows down towards the end since the last
trees tend to have only singleton itemsets as nodes.

\subsection{Reducing itemsets}

Our next goal was to study the sizes of the generated decomposable
families compared to the size of the original candidate set. As input
for this experiment, we used several different candidate collections of
frequent itemsets resulting from varying the support threshold, and
generated the corresponding decomposable itemset families
(Table~\ref{tab:frequent}).

\begin{table}[htb!]
\begin{center}
\begin{tabular}{l@{\hspace{0.2cm}}rr@{\hspace{0.4cm}}rrrr@{\hspace{0.4cm}}rrrr}
\toprule
& & & \multicolumn{4}{c}{First Family, $\abs{\cover{G}_1}$} & \multicolumn{4}{c}{All Families, $\abs{\bigcup \cover{G}_i}$} \\
\cmidrule{4-7}
\cmidrule{8-11}
Dataset & $\sigma$ & $\abs{\cover{F}}$ & AIC & BIC & MDL & None & AIC & BIC & MDL & None \\
\midrule
\dtname{Mammals} & $.20$ & $2169705$ & $221$ & $213$ & $215$ & $10663$ & $668$ & $625$ & $630$ & $11103$ \\
\dtname{Mammals} & $.25$ & $416939$ & $201$ & $197$ & $197$ & $6820$ & $535$ & $507$ & $509$ & $7106$ \\
\dtname{Paleo} & $.01$ & $22283$ & $339$ & $281$ & $290$ & $5260$ & $993$ & $834$ & $812$ & $6667$ \\
\dtname{Paleo} & $.02$ & $979$ & $254$ & $235$ & $239$ & $376$ & $463$ & $433$ & $429$ & $733$ \\
\dtname{Paleo} & $.03$ & $298$ & $191$ & $190$ & $190$ & $210$ & $231$ & $228$ & $228$ & $277$ \\
\dtname{Paleo} & $.05$ & $157$ & $147$ & $147$ & $147$ & $151$ & $149$ & $149$ & $149$ & $156$ \\
\dtname{Courses} & $.01$ & $16945$ & $217$ & $202$ & $206$ & $4087$ & $565$ & $522$ & $524$ & $4357$ \\
\dtname{Courses} & $.02$ & $2493$ & $185$ & $177$ & $177$ & $625$ & $354$ & $342$ & $342$ & $751$ \\
\dtname{Courses} & $.03$ & $773$ & $176$ & $170$ & $170$ & $276$ & $264$ & $261$ & $261$ & $359$ \\
\dtname{Courses} & $.05$ & $230$ & $136$ & $132$ & $132$ & $158$ & $167$ & $164$ & $164$ & $186$ \\
\bottomrule
\end{tabular}
\caption{Sizes of decomposable families for various datasets. The
second column is the minimum support threshold, the third column is
the number of the frequent itemsets in the candidate set. The columns
4--7 contain the size of the first result family and the columns 8--11
contain the size of the union of the result families.}
\end{center}
\label{tab:frequent}
\end{table}

From the results we see that the decomposable families are much
smaller compared to the original candidate set, as a large portion of
itemsets are pruned due to the running intersection property. The
regularizations AIC, BIC, MDL prune the results further. The pruning
is most effective when the candidate set is large.

\subsection{Boolean Queries}
\label{sec:experiments:query}
We conducted a series of boolean queries for \dtname{Paleo} and
\dtname{Courses} datasets.  For each dataset we pick randomly $1000$
queries of size $5$. We constructed a sequence of trees using BIC and
the unconstrained (min. support = 0) candidate set as input.  The
average computation time for a single query was $0.3s$. A portion
(abt. 10\%) of queries had to be discarded due to the numerical
instability of the linear program solver we used.

A query $Q$ for a decomposible family $\cover{G}_i$ produces a
frequency interval $\freq{Q; \cover{G}_i}$. We also computed the
frequency interval $\freq{Q; \cover{I}}$, where $\cover{I}$ is a
family containing nothing but singletons.  We studied the ratios $r(Q;
n) = \abs{\bigcap_1^n \freq{Q; \cover{G}_i}} / \abs{\freq{Q;
\cover{I}}}$ as a function of $n$, that is, the ratio between the
tightness of the bound using $n$ families and the singleton model.

\begin{figure}
\center
\subfigure[\dtname{Improved queries}]{\label{fig:queryres:a}\includegraphics[width=4cm]{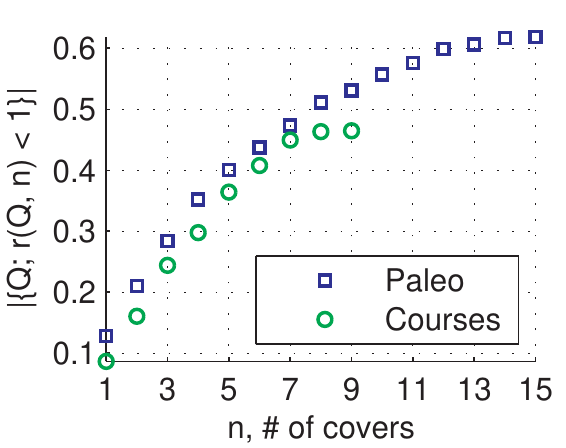}}%
\subfigure[\dtname{Paleo}]{\label{fig:queryres:b}\includegraphics[width=4cm]{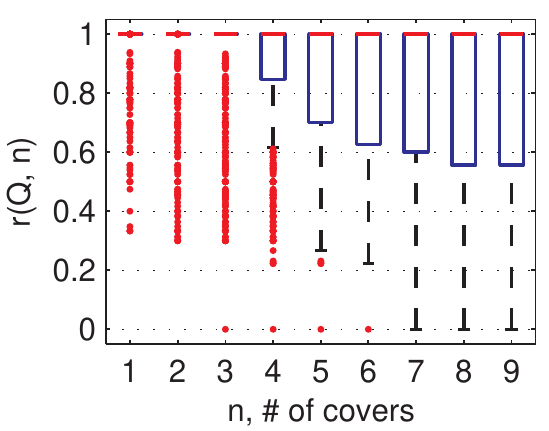}}%
\subfigure[\dtname{Courses}]{\label{fig:queryres:c}\includegraphics[width=4cm]{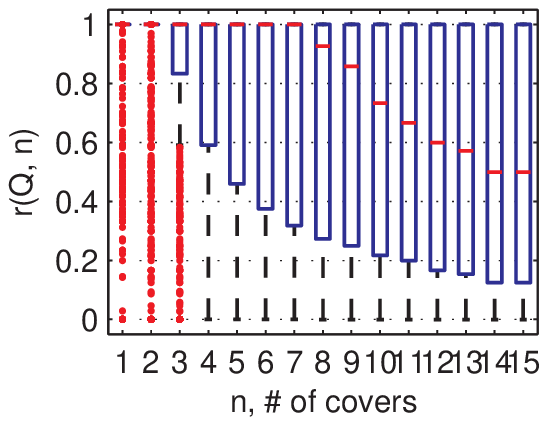}}

\caption{Boolean query ratios from \dtname{Paleo} and \dtname{Course}
datasets.  Figure~\ref{fig:queryres:a} contains the percentage of
queries having $r(Q; n) < 1$, that is, the percentage of queries
improved over the singleton model as a function of the number of
decomposable families.
Figures~\ref{fig:queryres:b}--\ref{fig:queryres:c} are box plots of
the ratios $r(Q; n)$, where $Q$ is a random query and $n$ is the
number of decomposable families.}

\label{fig:queryres}
\end{figure}

From the results given in Figure~\ref{fig:queryres} we see that the
first decomposible family in the sequence yields in about 10 \% of the
queries an improved bound with respect to the singleton family. As the
number of decomposable families increases, the number of queries with
tighter bounds goes from $10\%$ up to $60\%$. Also, in general the
absolute bounds become tighter for the queries as we increase the
number of decomposable families. For \dtname{Courses} the median of
the ratio $r(Q; 15)$ is about $0.5$.

\section{Related Work}
\label{sec:related}
One of the main uses of our algorithm is in reducing itemset mining
results into a smaller and a more manageable group of itemsets. One of
the earliest approaches on itemset reduction include close
itemsets~\cite{pasquier99discovering} and maximal frequent itemset
\cite{bayardo98}. Also more recently, a significant amount of
interesting research has been produced on the
topic~\cite{calders02mining,jiawei,siebesVL06,bringmannZ07}. Yan et
al. \cite{jiawei} proposed a statistical model in which $k$
representative patterns are used to summarize the original itemset
family as well as possible. This approach has, however, a different
goal to that of ours, as our model aims to describe the data
itself. From this point of view the work by Siebes et
al. \cite{siebesVL06} is perhaps the most in concordance to
ours. Siebes et al. propose an MDL based method where the reduced
group of itemsets aim to compress the data as well as possible. Yet,
their approach is technically and methodologically quite different and
does not provide a probabilistic model of the data as our model does.
Furthermore, non of the above approaches provide a naturally following
tree based representation of the mining results as our model does.

Traditionally, junction trees are not used as a direct model but
rather as a technique for decomposing directed acyclic graph (DAG)
models~\cite{cowell99network}. However, there is a clear difference
between the DAG models and our approach. Assume that we have 4 items
$a$, $b$, $c$, and $d$.  Consider a DAG model $p(a)p(b; a)p(c; a)p(d;
bc)$. While we can decompose this model using junction trees we cannot
express it exactly. The reason for this is that the DAG model contains
the assumption of independence of $b$ and $c$ given $a$. This allows
us to break the clique $abc$ into smaller parts.  In our approach the
cliques are the empirical distributions with no independence
assumptions. DAG models and junction tree models are equivalent for
Chow-Liu tree models~\cite{chow68treemodel}.

Our algorithm for constructing junction trees is closely related to EFS
algorithm~\cite{deshpande02junction,altmueller04efs} in which new cliques are
created in a similar fashion. The main difference between the approaches is
that we add new cliques in a level-wise fashion. This allows a more
straightforward algorithm. Another benefit of our approach is
Theorem~\ref{thr:coveroptimal}. On the other hand, Corollary~\ref{cor:chowliu}
implies that our algorithm can be seen also as an extension of Chow-Liu tree
model~\cite{chow68treemodel}.

\section{Conclusions and Future Work}
\label{sec:conclusions}
In this study we applied the concept of junction trees to create
decomposable families of itemsets. The approach suits well for the
problem of itemset selection, and has several advantages. The
naturally following junction trees provide an intuitive representation
of the mining results. From the computational point of view, the model
provides leverage for problems that could be intractable using generic
families of itemsets. We provided an efficient algorithm to build
decomposable itemset families, and gave an application example with
frequency bound querying using the model. Empirical results showed
that our algorithm yields high quality results. Because of the
expressiveness and good interpretability of the model, applications
such as classification using decomposable families of itemsets could
prove an interesting avenue for future research. Even more generally,
we anticipate that in the future decomposable models could prove
computationally useful with pattern mining applications that otherwise
could be hard to tackle.

\bibliography{partition}
\appendix
\section{Appendix}
\subsection*{Proof of Theorem~\ref{thr:junctionlegal}}
The theorem is trivial for the case $n = 1$. Hence we assume that $n > 1$.

Assume that $X$ and $Y$ are $n-1$-connected in $\tree{T}$, and let $P$ be the
path connecting $X$ and $Y$. If $\pr{X, Y} \notin P$ , then there exists
$f = \pr{Z, W} \in P$ such that $\abs{Z \cap W} = n - 1$. Since $\tree{T}$
is a junction tree, we must have $Z \cap W = X \cap Y$. Hence by removing $f$
and adding $\pr{X, Y}$ does not violate the running intersection property.
Thus we can assume that $\pr{X, Y} \in \edge{\tree{T}}$. Adding $V$ between
$X$ and $Y$ does not violate the running intersection property and hence
$\vrt{\tree{T}} + V$ remains decomposable.

To prove the other direction assume that $X$ and $Y$ are not $n-1$ connected.
This implies that the path $P$ from $X$ to $Y$ consists of $n + 1$ cliques with
$n$ separators. Assume that $\vrt{\tree{T}} + V$ is decomposable and hence
there is a juncion tree $\tree{U}$ having $\vrt{\tree{U}} = \vrt{\tree{T}} + V$.
We can modify $\tree{U}$ such that the edges of the path $P$ occur in $\tree{U}$.
Let $P_s$ be the first clique in $P$ and let $P_l$ be the last.
Note that $V \subset P_s \cup P_l$. Let $P_e$ be the first clique in $P$ along the
path from $V$ to $P_s$. Since $\abs{P_s \cap V} = n$, we must have $P_e \cap V = P_s \cap V$
The path from $V$ to $P_l$ must also go through $P_e$, hence we must have
$P_s \cap V = P_l \cap V$. This implies that either $V = P_s$ or $V = P_l$, which
is a contradiction. This completes the proof.

\subsection*{Proof of Theorem~\ref{thr:edgeweight}}
Assume that the edge $e = \pr{X, Y} \in \edge{\tree{T}_n}$. Let $\tree{T}'_n$
be the tree after adding $X \cup Y$.
The entropy of the original tree is
\[
\ent{\tree{T}_n} = \ent{X} + \ent{Y} - \ent{X \cap Y} + B,
\]
where $B$ is the impact of the rest nodes. The entropy of the new tree is
\[
\ent{\tree{T}'_n} = \ent{X \cup Y} + B.
\]
Hence we have $\ent{\tree{T}_n} - \ent{\tree{T}'_n} = w(e)$.

\subsection*{Proof of Theorem~\ref{thr:coveroptimal}}
It is easy to see that the cliques $X$ and $Y$ are $n-1$ -connected if and
only if they are not connected by the previous edges from $G_n$. Hence, the
algorithm reduces to Krus\-kal's algorithm in finding the optimal spanning
tree of $G_n$, thus returning the optimal spanning tree.

Let $\tree{U}$ be a junction tree refined by $\tree{T}_n$ and containing the
cliques of size $n + 1$, at maximum. The cliques of size $n + 1$ occur in $G_n$.
Let $H$ be the corresponding edges in $G_n$. To prove the theorem we need to
show that $H$ contains no cycles.

Assume othewise, and consider adding the edges in $H$, one at the time.
When the first cycle occurs, the corresponding family is not decomposable by
Theorem~\ref{thr:junctionlegal}. The argument in the proof of Theorem~\ref{thr:junctionlegal}
holds even if we keep adding cliques of size $n + 1$, hence the final family cannot
be decomposable. Thus $H$ cannot contain cycles.

\subsection*{Proof of Theorem~\ref{thr:query}}
Theorem 6 in~\cite{dobra00bounds} guarantees that 
breaking $\cover{G}$ into connected components and computing $\freq{Q; \cover{G}}$
from $\alpha_i$ and $\beta_i$ produce an accurate result as long as $\alpha_i$ and $\beta_i$
are accurate. Theorem 7 in~\cite{tatti06projections} states that
taking the smallest subtree containing $Q_i$ and removing attributes occuring in only
one clique does not change $\alpha_i$ and $\beta_i$. 

Finally, we need to prove that the linear program of the algorithm produce the
same $\alpha_i, \beta_i$ as the linear program in Eq.~\ref{eq:maxfreq}. Let $p$ be
a distribution satisfying the conditions in Eq.~\ref{eq:maxfreq}. Clearly, we can
break $p$ into components satisfying the conditions of the linear program given in
the algorithm. On other hand, assume that $\set{p_C}$ now satisfy the conditions of the
linear program given in the algorithm. Since components are equal at the separators
we can combine this into one joint distribution $p$ satisfying the condition of Eq.~\ref{eq:maxfreq}.
This implies that the outcome of both programs are equivalent.

To prove the bound for the number of variables, note that for any clique $C$ we
have $2^{\abs{C}} \leq \abs{\cover{G}}$. We can have $\abs{A}$ cliques at most.
Augmenting can increase the size of the cliques by $\abs{Q}$, at maximum. This
implies that the number of variables is $\sum_i 2^{\abs{Q} + \abs{C_i}} =
2^{\abs{Q}}\sum_i 2^{\abs{C_i}} \leq 2^{\abs{Q}}\abs{\cover{G}}\abs{A}$.

\begin{figure}[p!]
\includegraphics[width=\textwidth, angle=-90]{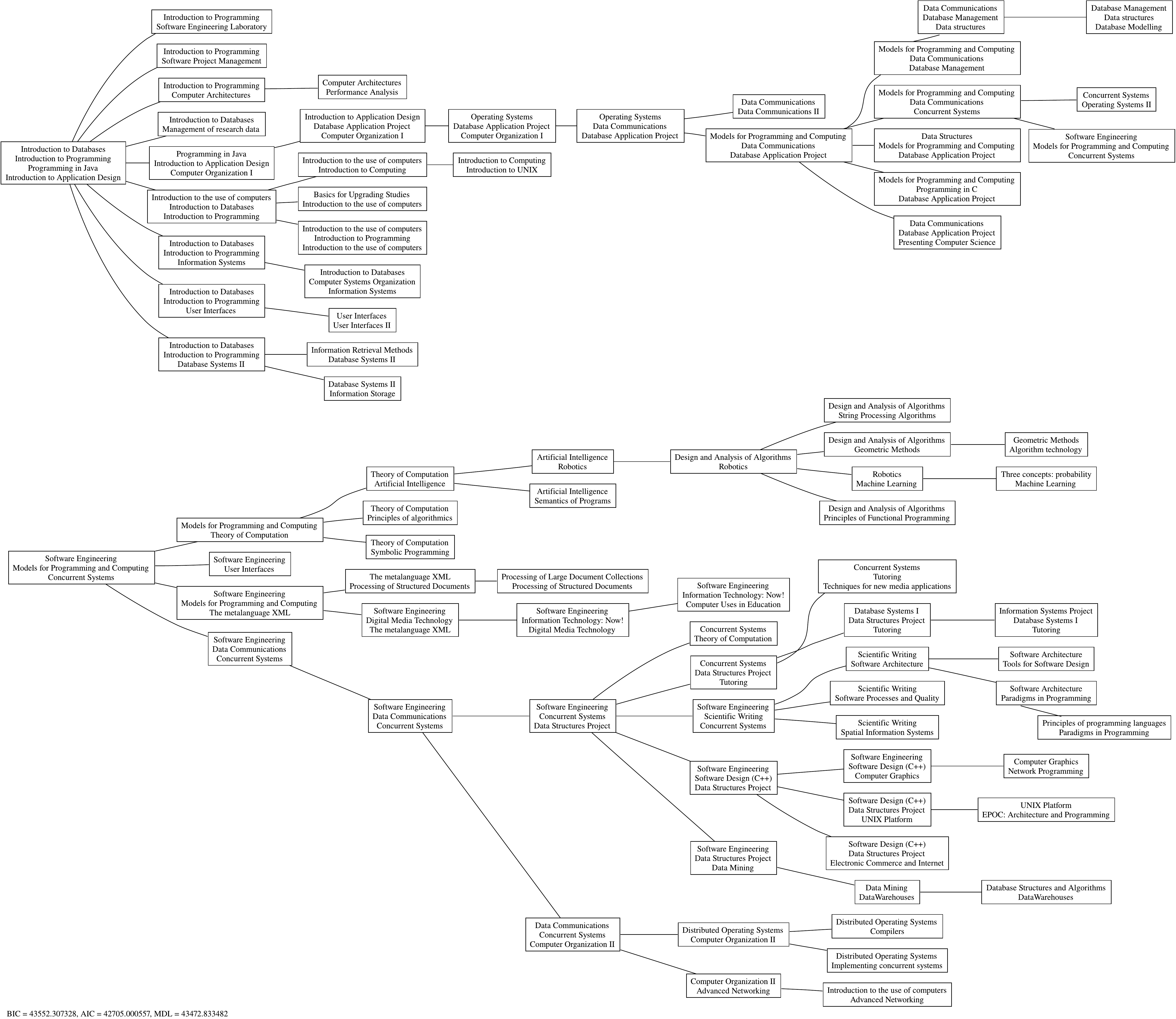}
\caption{Junction tree build from the \dtname{Courses} dataset. The
tree was constructed using an unconstrained candidate family
(min. support = 0) as input and BIC as regularization.}
\end{figure}

\end{document}